\documentclass[runningheads,a4paper]{llncs}

\usepackage{array}
\usepackage{amssymb}
\usepackage{graphicx}
\usepackage{subfig}
\usepackage{float}
\usepackage{placeins}
\usepackage{url}
\usepackage{tabularx}
\usepackage{rotating}
\usepackage[utf8]{inputenc}

\usepackage{blindtext}
\usepackage{everypage}
\usepackage{environ}
\usepackage{pdflscape}

\usepackage[colorinlistoftodos]{todonotes}

\newcounter{abspage}
\usepackage{atbegshi}
\AtBeginDocument{\AtBeginShipoutNext{\AtBeginShipoutDiscard}}

\makeatletter
\newcommand{\newSFPage}[1]
  {\global\expandafter\let\csname SFPage@#1\endcsname\null}

\NewEnviron{SidewaysFigure}{\begin{figure}[p]
\protected@write\@auxout{\let\theabspage=\relax}
  {\string\newSFPage{\theabspage}}%
\ifdim\textwidth=\textheight
  \rotatebox{90}{\parbox[c][\textwidth][c]{\linewidth}{\BODY}}%
\else
  \rotatebox{90}{\parbox[c][\textwidth][c]{\textheight}{\BODY}}%
\fi
\end{figure}}

\AddEverypageHook{
  \ifdim\textwidth=\textheight
    \stepcounter{abspage}
  \else
    \@ifundefined{SFPage@\theabspage}{}{\global\pdfpageattr{/Rotate 0}}%
    \stepcounter{abspage}%
    \@ifundefined{SFPage@\theabspage}{}{\global\pdfpageattr{/Rotate 90}}%
  \fi}
\makeatother

\urldef{\mailsa}\path|{adria.casamitjana,veronica.vilaplana}@upc.edu|

\begin{document}


\title{Cascaded V-Net using ROI masks for brain tumor segmentation}

\titlerunning{Cascaded V-Net using ROI masks}

%
%
\author{Adri\`a Casamitjana, Marcel Cat\`a, Irina S\'anchez, Marc Combalia and Ver\'onica Vilaplana }%
\thanks{This work has been partially supported by the projects BIGGRAPH-TEC2013-43935-R and MALEGRA TEC2016-75976-R financed by the Spanish Ministerio de Econom\'{i}a y Competitividad and the European Regional Development Fund (ERDF). Adri\`{a} Casamitjana is supported by the Spanish “Ministerio de Educación, Cultura y Deporte” FPU Research Fellowship. }%
%
\authorrunning{Adri\`a Casamitjana et al. }

\institute{Signal Theory and Communications Department, Universitat Polit\`{e}cnica de Catalunya. BarcelonaTech, Spain\\
\mailsa\\}

%
%

\toctitle{Lecture Notes in Computer Science}
\tocauthor{Authors' Instructions}
\maketitle
\begin{abstract}
\setcounter{page}{1}
In this work we approach the brain tumor segmentation problem with a cascade of two CNNs inspired in the V-Net architecture \cite{VNet}, reformulating residual connections and making use of ROI masks to constrain the networks to train only on relevant voxels. This architecture allows dense training on problems with highly skewed class distributions, such as brain tumor segmentation, by focusing training only on the vecinity of the tumor area. We report results on BraTS2017 Training and Validation sets.
\end{abstract}

\section{Introduction}

Accurate localization and segmentation of brain tumors in Magnetic Resonance Imaging (MRI) is crucial for monitoring progression, surgery or radiotherapy planning and follow-up studies. Since manual segmentation is time-consuming and may lead to inter-rater discrepancy, automatic or semi-automatic approaches have been a topic of interest during the last decade.
Among tumors that originally develop in the brain, gliomas are the most common type. Gliomas may have different degrees of aggressiveness, variable prognosis and several heterogeneous histological sub-regions (peritumoral edema, necrotic core, enhancing and non-enhancing tumor core) that are described by varying intensity profiles across different MRI modalities, which reflect diverse tumor biological properties \cite{Brats15}. However, the distinction between tumor and normal tissue is difficult as tumor borders are often fuzzy and there is a high variability in shape, location and extent across patients. Despite recent advances in automated algorithms for brain tumor segmentation in multimodal MRI scans, the problem is still a challenging task in medical imaging analysis.

Many different computational methods have been proposed to solve the problem. 
Here we will only review some of the most recent approaches based on deep learning, which are the top-performing methods in BraTS challenge since 2014. Representative works based on other machine learning models include \cite{Gooya,Zikic,Parisot,Maier,Tustison} and methods reviewed in \cite{Brats15}.

As opposed to classical discriminative models based on pre-defined features, deep learning models learn a hierarchy of increasingly complex task specific features directly from data, which results in more robust features. 

Some methods do not completely exploit the available volumetric information and use two-dimensional Convolutional Neural Networks (CNN), processing 2D slices independently or using three orthogonal 2D patches to incorporate contextual information \cite{Pereira,Havaei}. The model in \cite{Havaei} consists of two pathways, a local pathway that concentrates on pixel neighborhood information, and a global pathway, which captures global context of the slice. This two-path structure is adopted in a fully 3D approach named DeepMedic \cite{Kamnitsas1}, consisting of two parallel 3D CNN pathways producing soft segmentation maps, followed by a fully connected 3D CRF that imposes generalization constraints. The network is extended in \cite{Kamnitsas2} by adding residual connections between the outputs of every two layers. The work shows empirically that the residual connections give modest but consistent improvement in sensitivity over all tumor classes. In \cite{Casamitjana} we compare the performances of three 3D CNN architectures inspired in two well known 2D models used for image segmentation \cite{Long,UNet} and a variant of \cite{Kamnitsas1} showing the importance of the multi-resolution connections to obtain fine details in the segmentation of tumor sub-regions. More recently, V-Net \cite{VNet} presents successful results on challenging medical imaging segmentation tasks by using both short and long skip connections that help learning finer structures and ease training.

In this paper, in the context of BraTS Challenge 2017, we present a  brain tumor segmentation method based on a cascade of two convolutional neural networks. The problem is divided in two simpler tasks that can be performed independently using two 3D-CNN and a later combination of their outputs to get the final segmentation.
The network architecture used for the two tasks is a modified version of V-Net consisting of convolutional blocks and residual connections that have been reformulated according to recent findings in the literature \cite{He:id_mappings}. Additionally, we introduce the use of ROI masks during the learning process in order to constrain each CNN to focus only on relevant voxels or regions from each task. Hence, the first network will be trained only on brain tissue to produce raw tumor masks and the second networks will be trained on the vecinity of the tumor to predict tumor regions.

Medical images in general and brain MRI in particular contain non-informative voxels (e.g. background or non-brain tissue) and many techniques have been developed to filter out this information. We can benefit from this knowledge and focus the entire system to train only on relevant, informative voxels or regions (e.g. apply a skull-stripping method and work with the brain mask to discard background information). To do so, the loss is computed only within the mask and the outer voxels will not contribute in the learning process, blocking the backpropagated signal through them. Finally, we use a dense-training scheme with small batch sizes that avoids patch-wise training and reduces the overall training time. Moreover, the common structure of the brain across subjects may be better learned using the whole image for training.



\section{Method}
\begin{figure}
\centering
\includegraphics[width=\textwidth]{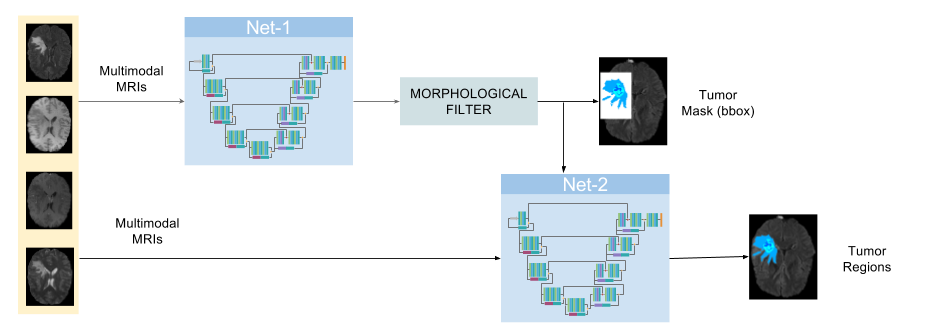}
\caption{The pipeline used for brain tumor segmentation}
\label{fig:pipeline}
\end{figure}

One of the main problems in brain lesion detection is that lesions affect a small portion of the brain, making naive training strategies biased towards the trivial decision of null detection. Brain tumors normally correspond to only 3-5\% of the overall image, accounting for 5-15\% of the brain tissue and being each tumor region an even smaller portion. To address this issue, we propose to divide the brain tumor segmentation problem into two simpler tasks: (i) segmentation of the overall tumor and (ii) delineation of the different tumor regions. The tasks are performed in parallel using two CNN networks, where the output of the first network is used an input to the second one. The overall system pipeline is depicted in Fig.~\ref{fig:pipeline}.

\subsection{V-Net using ROI masks}
\begin{figure}[h!]
\begin{tabular}{cc}
\subfloat[]{\includegraphics[width = 1.5in,height = 1.6in]{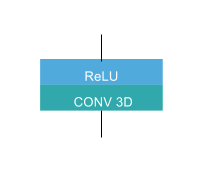}} &
\subfloat[]{\includegraphics[width = 1.5in,height = 1.6in]{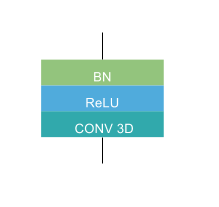}} \\
\subfloat[]{\includegraphics[width = 1.5in,height = 1.6in]{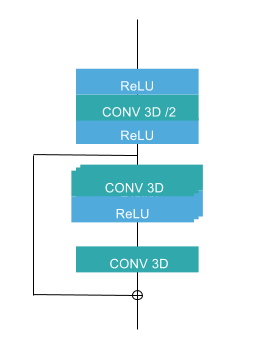}} &
\subfloat[]{\includegraphics[width = 1.5in,height = 1.6in]{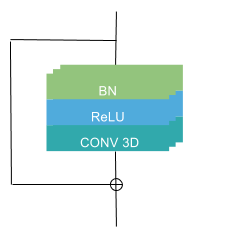}}
\end{tabular}
\centering

\caption{Comparison between original V-Net modules and proposed modifications. First row: basic convolutional block of V-Net (a) and corresponding modification using BN (b). Second row: standard V-Net residual connections (c) and proposed residual connections (d).}

\label{fig:vnet_blocks_mod}
\centering
\end{figure}

Our network is a variant of V-Net \cite{VNet} that aims at reducing the overall number of parameters by using smaller filter sizes (3x3x3 instead of 5x5x5) and changing the non-linearity from PReLU to ReLU. In addition, we use batch normalization before the non-linearity to account for internal covariate shift. Based on insights from \cite{He:id_mappings}, we also reformulate the short residual connections in order to improve gradients flow across the network by using identity mappings as residual connections. In the case of dimensions mismatch in the addition layer, we minimally modify the residual connection with max-pooling and repeated up-sampling for spatial correspondence and 1x1x1 convolutions to match the number of channels. For better understanding, we show in Fig.~\ref{fig:vnet_blocks_mod} the main changes from the original V-Net. 

\begin{sidewaysfigure}
\centering
\includegraphics[width=\textwidth]{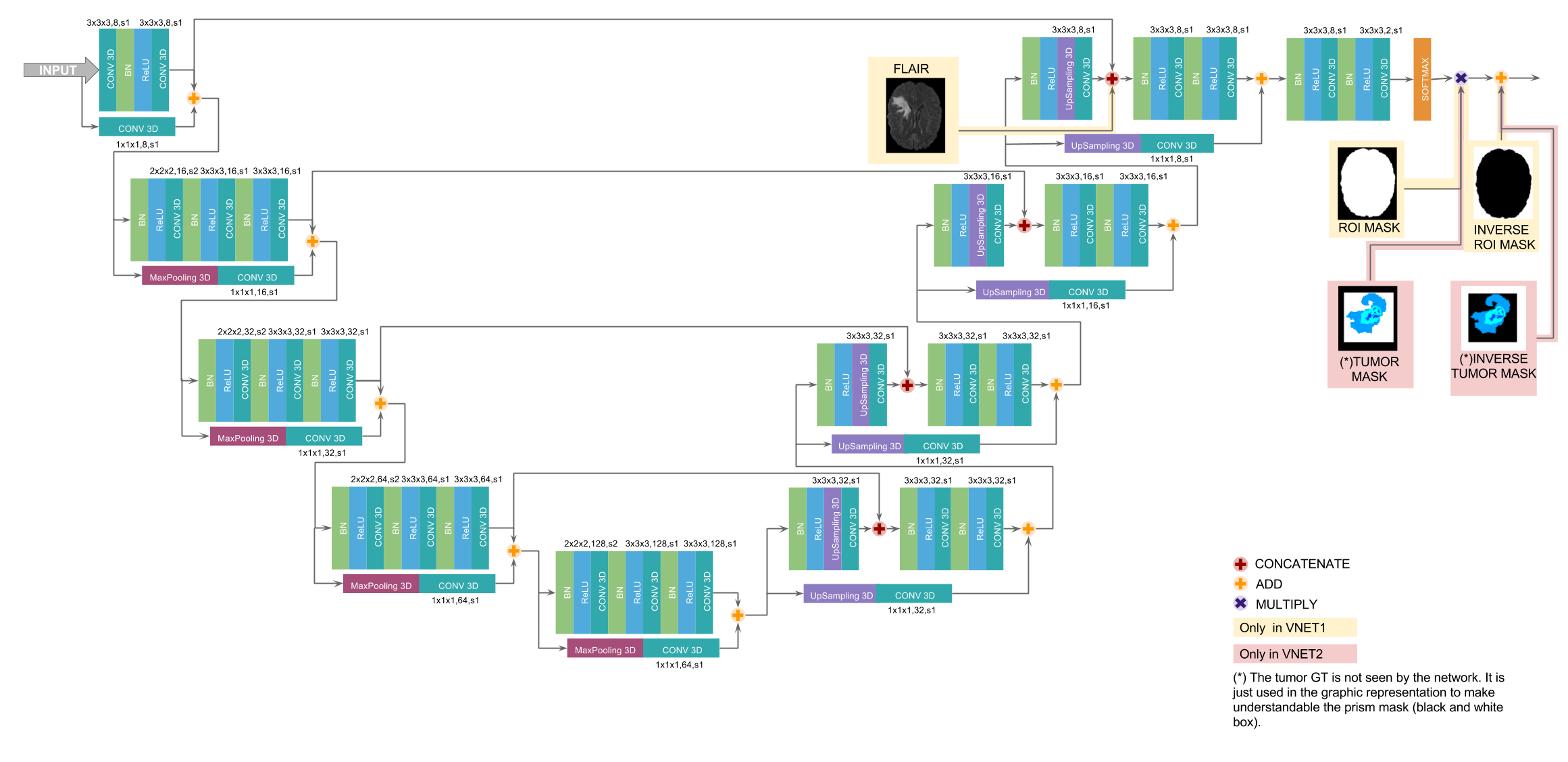}
\caption{V-Net architecture used in this work}
\label{fig:vnet-mask}
\end{sidewaysfigure}

We use ROI masks before the final predictions both during training and inference in order to smooth the class imbalance problem, specially for small sub-tumor regions. The ROI mask forces the outer voxels to belong to the background class with full probability by first multiplying them by 0 and turn their probability of belonging to the background class to 1. The multiplication prevents the backpropagated signal from going through the outer voxels and thus, it does not contribute in the learning process. The overall architecture is shown in Fig.~\ref{fig:vnet-mask}. 


\subsection{Training}
Each network in the pipeline performs a different task and thus, they can be independently trained. Instead of using patch-wise training and non-uniform sampling strategies to account for class imbalance, we use dense-training with a single subject per batch.
The first network is trained as a binary segmentation problem with tumor/non-tumor classes and outputs a raw segmentation of the whole-tumor region. It takes the four modalities (T1, T1c, T2 and FLAIR) as inputs and uses the FLAIR intensity information in the deeper layers of the network by concatenating it with the predicted feature maps from the last level in the expanding path of the network. The network  makes use of a brain mask in order to consider only brain tissue voxels for training. The loss function used is the modified dice coefficient (\ref{eq:dice_loss}) suited for binary segmentation tasks with imbalanced data:

\begin{equation}
L_1=\frac{\sum_{i=1}^{N} p_i \cdot l_i}{\sum_{i=1}^{N} p_i +  \sum_{i=1}^{N} l_i }
\label{eq:dice_loss}
\end{equation}

where N is the total number of voxels, $p_i$ is the softmax output of the $i$-th voxel, and $l_i$ is the $i$-th voxel label ($l_i = 0,1$).

The second network is trained as a multi-class segmentation problem with four classes (non-tumor, edema, enhancing core and non-enhancing core). It also uses the four MRI modalities as input. From the ground-truth labels, we generate a rectangular mask that covers the whole tumor and it is used to train the network only in the vicinity of the tumor, avoiding to train on brain tissue far from the tumor region. As we use dense-training with a single subject per batch, the use of raw tumor masks in the training procedure helps to reduce the  high class imbalance present among different classes. The loss function in the second network is a combination of cross entropy ($X_E$) and the dice coefficient for each tumor sub-region (whole tumor ($D_{WT}$), enhancing tumor ($D_{ET}$) and tumor core ($D_{TC}$)). We empirically choose the values for the weights in both parts of the function:

\begin{equation}
L_2= X_E + 0.5 * ( {D_{WT} + D_{ET} + D_{TC}})
\end{equation}

\subsection{Inference}

At inference time, we first get the whole-tumor prediction from the first network. We use morphological filtering to remove small spurious detections made by the first network and we automatically find the smallest rectangular mask that covers the detected tumor. This result is then used as ROI mask in the second network to mask out the majority of false positives, since the second network is only trained to discriminate tumor regions on the vicinity of the tumor. This process can be done in parallel since one can decouple the masking process from the network prediction and thus, save inference time.

\section{Results and discussion}

\subsection{Data} 
BraTS2017 training data \cite{Data1,Data2,Data3} consists of 210 pre-operative MRI scans of subjects with glioblastoma (GBM/HGG) and  75 scans of subjects with lower grade glioma (LGG), corresponding to the following modalities: native T1, post-contrast T1-weighted, T2-weighted and FLAIR, acquired from multiple institutions. Ground truth annotations comprise GD-enhancing tumor (ET, label 4), peritumoral edema (ED, label 2), necrotic and non-enhancing tumor (CNR/NET, label 1) as described in \cite{Brats15}. The data is distributed co-registered to the same anatomical template, interpolated to the same resolution ($1mm^3$) and skull-stripped. The validation set consists of 46 scans with no distinction between GBM/HGG and LGG.

Each scan is individually normalized in mean and standard deviation. For training, we use data augmentation by adding scan reflections with respect to the sagittal plane.

\subsection{Performance on BraTS2017 training and validation sets}


Evaluation of the results is performed merging the predicted labels into three classes: enhancing tumor ET (label 1), whole tumor WT (labels 1, 2, 4), and tumor core TC (labels 1, 4), using Dice score, Hausdorff distance, Sensitivity and Specificity. Preliminary results for the BraTS 2017 Training dataset have been obtained by hold-out using 70\% of the data for training and the remaining 30\% for development purposes, such as early stopping or to tune some hyperparameter. In addition to that, the performance on the BraTS 2017 Validation set, reported on the challenge's leaderboard \footnote{https://www.cbica.upenn.edu/BraTS17/lboardValidation.html}, is also presented in Table \ref{tab:results_performance_1} and Table \ref{tab:results_performance_2}.

In Table \ref{tab:results_performance_1} we show Dice and Hausdorff metrics. We achieve rather high performance on the Dice metric for the whole tumor (WT) region, but low values for enhancing tumor (ET) and tumor core (TC) regions, compared to state-of-the-art.
Using the BraTS Validation set, we are able to compare to other participants in the challenge. In the case of Dice-WT, our method is very close to the results obtained by the top performing methods while, again, our method achieves rather low Dice-ET and Dice-TC metrics. Hausdorff distances are higher than the best performing algorithms, being specially high for the whole-tumor region, probably indicating some outlier predictions that increase the metric.

\begin{table}[h]
\setlength\tabcolsep{3.5mm}
\centering
\begin{tabular}{ c c c c | c c c }
\hline
&\multicolumn{3}{c}{Dice}  &\multicolumn{3}{c}{Hausdorff}\\
\hline 
& ET & WT & TC & ET & WT & TC  \\
\hline
 Development set & 0.671 & 0.869 & 0.685 & 7.145 & 6.410 & 9.584\\
 Validation set & 0.714 & 0.877 & 0.637 & 5.434 & 8.343 & 11.173\\
\hline  
\end{tabular}
\caption{Results for BraTS 2017 data. Dice and Hausdorff metrics are reported.}
\label{tab:results_performance_1}
\centering
\end{table}

Even though specificity is not very informative for imbalanced classes, results from Table~\ref{tab:results_performance_2} show that we are able to properly represent background, probably due to the use of masks in the predictions. More interestingly, sensitivity shows that ET and TC regions might be underrepresented in our predicted segmentations. This results guide us to future improvements trying to overcome that behavior.

\begin{table}[h]
\setlength\tabcolsep{3.5mm}
\centering
\begin{tabular}{ c c c c | c c c }
\hline
&\multicolumn{3}{c}{Sensitivity} &\multicolumn{3}{c}{Specificity} \\
\hline 
& ET & WT & TC & ET & WT & TC \\
\hline
 Development set & 0.735 & 0.851 & 0.664 & 0.998 & 0.994& 0.997 \\
 Validation set & 0.723 & 0.879 & 0.619 & 0.998 & 0.994 & 0.998 \\
\hline  
\end{tabular}
\caption{Results for BraTS 2017 data. Sensitivity and specificity are reported.}
\label{tab:results_performance_2}
\centering
\end{table}

\begin{figure}[h!]

\begin{tabular}{cccc}
\subfloat[]{\includegraphics[width = 1.5in,height = 1.6in]{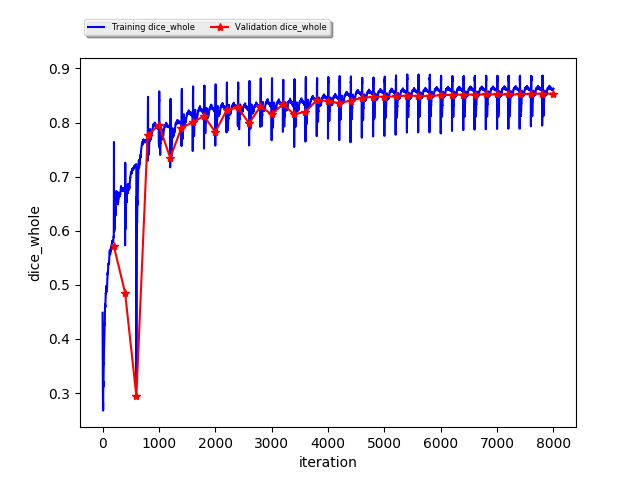}} &
\subfloat[]{\includegraphics[width = 1.5in,height = 1.6in]{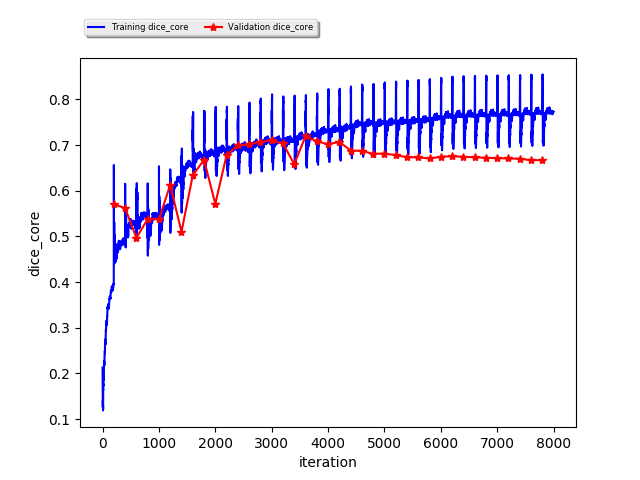}}&
\subfloat[]{\includegraphics[width = 1.5in,height = 1.6in]{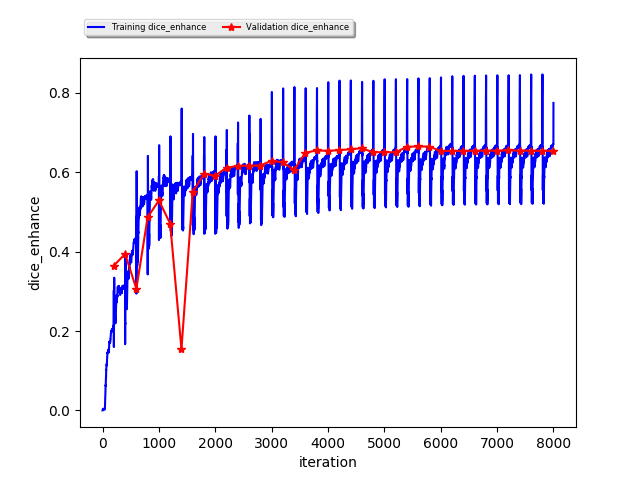}} 
\end{tabular}
\centering
\caption{Evaluation of metrics of interest during training in  training (blue) and development (red) sets. Metrics: dice whole tumor (a), dice tumor core (b) and dice enhancing tumor (c)}
\centering
\label{fig:train_dev_plots}
\end{figure}

These results can be further analyzed and confirmed with Train/Development curves in Fig.~\ref{fig:train_dev_plots}, which mostly indicate the generalization power of the method by analyzing the bias/variance trade-off. We clearly see that from epoch 18 (iteration 3600) we start overfitting the tumor core metric (Fig.~\ref{fig:train_dev_plots}.b) while the other regions are either slowly gaining insignificant improvements (Fig.~\ref{fig:train_dev_plots}.a, whole tumor) or not improving at all (Fig.~\ref{fig:train_dev_plots}.c, enhancing core). However the development loss is still improving due to the cross-entropy term. 

\subsection{Visual analysis}
\begin{figure}[h!]
\begin{tabular}{ccc}
{\includegraphics[width = 1.5in,height = 1.6in]{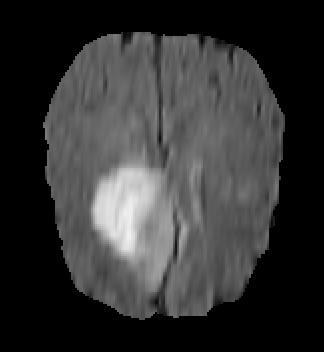}} &
\subfloat[]{\includegraphics[width = 1.5in,height = 1.6in]{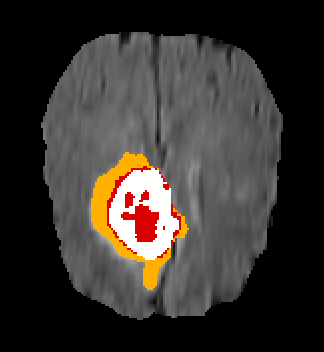}}&
{\includegraphics[width = 1.5in,height = 1.6in]{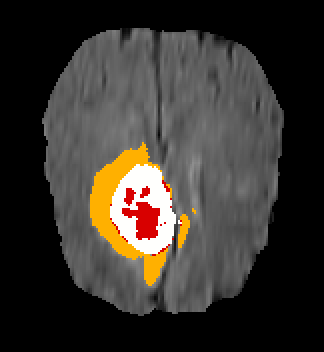}} \\
{\includegraphics[width = 1.5in,height = 1.6in]{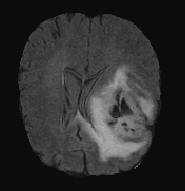}} &
\subfloat[]{\includegraphics[width = 1.5in,height = 1.6in]{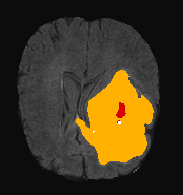}}&
{\includegraphics[width = 1.5in,height = 1.6in]{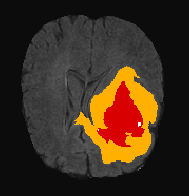}} 
\end{tabular}
\centering
\caption{Segmentation results of two subjects: a) TCIA 479 b) TCIA 109. From left to right we show the FLAIR sequence, followed by Prediction and GT tumor segmentation. We distinguish intra-tumoral regions by color-code: enhancing tumor (white), peritumoral edema (orange) and necrotic and non-enhancing tumor (red).}
\centering
\label{fig:subjects_comparison}
\end{figure}

Figure \ref{fig:subjects_comparison} shows two subjects among the quantitatively better (first row) and poorer (second row) results. In both cases, it can be visually appreciated that our method correctly segments the whole tumor region.
For the subject shown in Figure \ref{fig:subjects_comparison}.a, the system is able to properly capture all tumor regions, meaning that the first network is able to correctly localize the tumor and the second network is able to capture differences between tumor regions. On the other hand, in Figure \ref{fig:subjects_comparison}.b, we show a case where even though the tumor is correctly localized by the first network, the second isn't able to properly detect different tumor subregions. We see that edema (ED - label 2) is overrepresented in our segmentation in detriment of smaller classes: GD-enhancing tumor (ET - label 4) and the necrotic and non-enhancing tumor (NCR/NET - label 1). This effect can also be inferred from lower values in ET and TC dice coefficients.

\section{Conclusions}
In this paper we introduce a cascaded V-Net architecture that uses masks to focus training on relevant parts of the brain. We use it to solve the class imbalance problem inherent to brain tumor segmentation. We use a two-step process that (i) localizes the brain tumor area and (ii) distinguishes between different tumor regions, ignoring all other background voxels. This scheme allows to perform dense-training on MR images. We finally show results on BraTS 2017 Training and Validation sets, showing that while the results obtained for the WT segmentation are competitive with other participants' algorithms, we are not able to properly capture the less common regions (TC or ET). When trying to avoid patch-wise sampling strategies and make use of dense-training scheme, smaller classes are not well detected even using ROI masks, meaning that more weight should be placed when learning those classes. As a future work we will explore how to up-weight small tumor regions in the learning procedure.

\end{document}